\documentclass[]{llncs}
\usepackage{graphicx}
\usepackage{comment}
\usepackage{amsmath,amssymb} % define this before the line numbering.
\usepackage{acronym}
\usepackage{color}
\usepackage[resetlabels,labeled]{multibib}
\usepackage[symbol]{footmisc}

\usepackage[width=122mm,left=12mm,paperwidth=146mm,height=193mm,top=12mm,paperheight=217mm]{geometry}

\usepackage{tabularx, lipsum}
\usepackage{hyperref}
\hypersetup{
    colorlinks=true,
    linkcolor=blue,
    filecolor=blue,      
    urlcolor=blue,
}

\usepackage{xr}
\makeatletter
\newcommand*{\addFileDependency}[1]{% argument=file name and extension
  \typeout{(#1)}
  \@addtofilelist{#1}
  \IfFileExists{#1}{}{\typeout{No file #1.}}
}
\makeatother

\definecolor{orange}{RGB}{255,127,0}
\definecolor{green}{RGB}{0,127,0}

\newcommand{\secref}[1]{Section \ref{#1}}
\newcommand{\figref}[1]{Figure \ref{#1}}
\renewcommand{\eqref}[1]{Equation \ref{#1}}

\newcommand{\ignore}[1]{}

\makeatletter
\newcommand{\printfnsymbol}[1]{%
  \textsuperscript{\@fnsymbol{#1}}%
}
\makeatother

\makeatletter
\@addtoreset{footnote}{section}
\makeatother

\begin{document}

\pagestyle{headings}
\mainmatter
\def\ECCVSubNumber{2481}  % Insert your submission number here

\author{Aviv Shamsian\thanks{equal contribution}\inst{1} \quad 
Ofri Kleinfeld\printfnsymbol{1} \inst{1} \quad 
Amir Globerson\inst{2} \quad
Gal Chechik\inst{1,3}}

\title{Learning Object Permanence from Video} % Replace with your title

%\begin{comment}
% \titlerunning{ECCV-20 submission ID \ECCVSubNumber} 
% \authorrunning{ECCV-20 submission ID \ECCVSubNumber} 

%\end{comment}
%******************


% \titlerunning{Learning Object Permanence from Video}
\institute{Bar-Ilan University, Ramat-Gan, Israel \and
Tel Aviv University, Tel Aviv, Israel \and
% \email{Gal.Chechik@biu.ac.il}\\
% \url{http://chechiklab.biu.ac.il} \and
NVIDIA Research, Tel-Aviv, Israel 
\url{https://chechiklab.biu.ac.il/~avivshamsian/OP/OP_HTML.html}}

%******************

\maketitle
\begin{abstract}
\emph{Object Permanence} allows people to reason about the location of non-visible objects, by understanding that they continue to exist even when not perceived directly. Object Permanence is critical for building a model of the world, since objects in natural visual scenes dynamically occlude and contain each-other. Intensive studies in developmental psychology suggest that object permanence is a challenging task that is learned through extensive experience.

Here we introduce the setup of learning Object Permanence from labeled videos. We explain why this learning problem should be dissected into four components, where objects are (1) visible, (2) occluded, (3) contained by another object and (4) carried by a containing object.  
The fourth subtask, where a target object is carried by a containing object, is particularly challenging because it requires a system to reason about a moving location of an invisible object. We then present a unified deep architecture that learns to predict object location under these four scenarios. We evaluate the architecture and system on a new dataset based on CATER, and find that it outperforms previous localization methods and various baselines.  

%\keywords{Object Permanence, Reasoning, Video Analysis}
\end{abstract}

\section{Introduction}
Understanding dynamic natural scenes is often challenged by objects that contain or occlude each other. To reason correctly about such visual scenes, systems need to develop a sense of \textit{Object Permanence} (OP) \cite{Piaget1954TheCO}. Namely, the understanding that objects continue to exist and preserve their physical characteristics, even if they are not perceived directly. For example, we want systems to learn that a pedestrian occluded by a truck may emerge from its other side, but that a person entering a car would ``disappear" from the scene. 

The concept of OP received substantial attention in the cognitive development literature. Piaget hypothesized that infants develop OP relatively late (at two years of age), suggesting that it is a challenging task that requires deep modelling of the world based on sensory-motor interaction with objects. Later evidence showed that children learn OP for occluded targets early \cite{aguiar1999,baillargeon1991object}. Still, only at a later age do children develop understanding of objects that are contained by other objects \cite{smitsman2009significance}. Based on these experiments we hypothesize that reasoning about the location of non-visible objects may be much harder when they are carried inside other moving objects.

\begin{figure}[t]
    \hspace{60pt}(a)\hspace{55pt}(b)\hspace{70pt}(c)\hspace{60pt}(d)
    \begin{center}
    \includegraphics[width=0.89\textwidth]{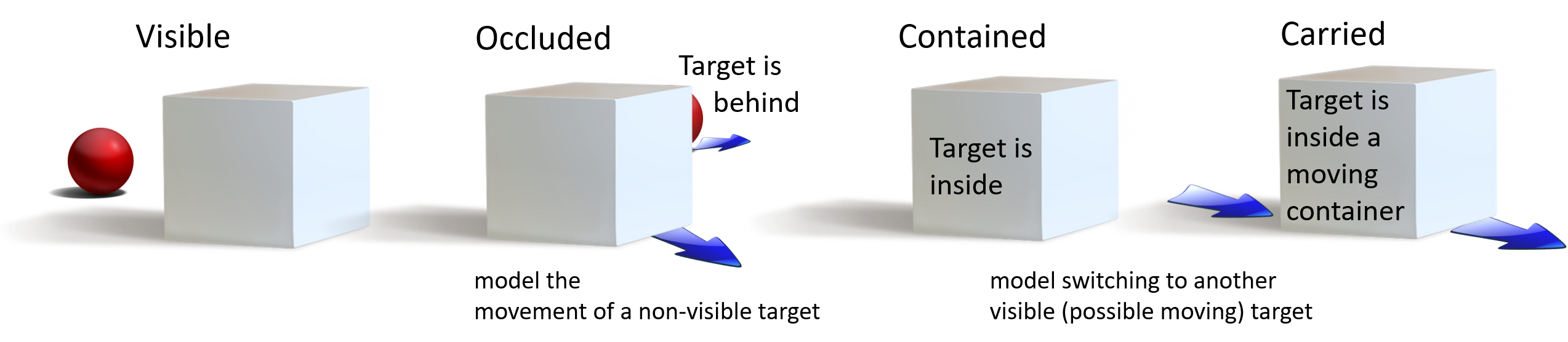}
    \caption{Inferring object location in rich dynamic scenes involves four different tasks, and two different types of reasoning. (a) The target, a red ball, is fully visible. (b) The target is fully-or partially occluded by the static cube. (c) The target is located inside the cube and fully covered. (d) The non-visible target is located inside another moving object; its location changes even though it is not directly visible \label{fig:object_per}}
    \end{center}
\end{figure}

Reasoning about the location of a target object in a video scene involves four different subtasks of increasing complexity (\figref{fig:object_per}). These four tasks are based on the state of the target object, depending if it is  (1) visible, (2) occluded, (3) contained or (4) carried. The \textit{visible} case is perhaps the simplest task, and corresponds to object detection, where one aims to localize an object that is visible. Detection was studied extensively and is viewed as a key component in computer vision systems.
The second task, \textit{occlusion}, is to detect a target object which becomes transiently invisible by a moving occluding object (e.g., bicycle behind a truck). Tracking objects under occlusion can be very challenging, especially with long-term occlusions \cite{mojtaba2019deep,Kristan2018TheSV,fan2019lasot,wu2015object}.

Third, in a \textit{containment} scenario, a target object may be located inside another container object and become non-visible \cite{ullman2019model} (e.g., a person enters a store).
Finally, the fourth case of a \textit{carried} object is arguably the most challenging task. It requires inferring the location of a non-visible object located inside a moving containing object (e.g., a person enters a taxi that leaves the scene). Among the challenging aspects of this task is the need to keep a representation of which object should be tracked at every time point and the need to ``switch states" dynamically through time. This task received little attention in the computer vision community so far.  

We argue that reasoning about the location of a non-visible object should address two distinct and fundamentally different cases: occlusion and containment. First, to \textit{localize an occluded object}, an agent has to build an internal state that models how the object moves. For example, when we observe a person walking in the street, we can predict her ever-changing location even if occluded by a large bus. In this mode, our reasoning mechanism keeps attending to the person and keeps inferring her location from past data. Second, \textit{localizing contained objects} is fundamentally different. It requires a reasoning mechanism that switches to attend to the containing object, which is visible. Here, even though the object of interest is not-visible, its location can be accurately inferred from the location of the visible containing object. We demonstrate below that incorporating these two reasoning mechanisms leads to more accurate localization in all four subtasks.

Specifically, we develop a unified approach for learning all four object localization subtasks in video. We design a deep architecture that learns to localize objects that may be visible, occluded, contained or carried. Our architecture consists of two reasoning modules designed to reason about (1) carried or contained targets, and (2) occluded or visible targets. The first reasoning component is explicitly designed to answer the question \textit{``Which object should be tracked now?"}. It does so by using an LSTM to weight the perceived locations of the objects in the scene. The second reasoning component leverages the information about which object should be tracked and previous known locations of the target to localize the target, even if it is occluded. Finally, we also introduce a dataset that is based on videos from CATER \cite{girdhar2019cater}, enriched with new annotations about task type and about ground-truth location of all objects.

Our main novel contributions are: (1) We conceptualize that localizing non-visible objects requires two types of reasoning: about occluded objects and about carried ones. (2) We define four subtypes of localization tasks and introduce annotations for the CATER dataset to facilitate evaluating each of these subtasks. (3) We describe a new unified architecture for all four subtasks, which can capture the two types of reasoning, and we show empirically that it outperforms multiple strong baselines.

\section{Related Work}

\subsection{Relational Reasoning in Synthetic Video Datasets} Recently, several studies provided synthetic datasets to explore object interaction and reasoning. Many of these studies are based on CLEVR \cite{johnson2017clevr}, a synthetic dataset designed for visual reasoning through visual question answering. CLEVRER \cite{yi2019clevrer} extended CLEVR to video, focusing on the causal structures underlying object interactions. It demonstrated that visual reasoning models that thrive on perception based tasks often perform poorly in causal reasoning tasks.

Most relevant for our paper, CATER \cite{girdhar2019cater} is a dataset for reasoning about object action and interactions in video. One of the three tasks defined in CATER, the \textit{snitch localization} task, is closely related to the OP problem studied here. It is defined as localizing a target \textit{at the end  of a video}, where the target is usually visible. 
Our work refines their setup, learning to localize the target through the full video, and breaks down prediction into four types of localization tasks. As a result, we provide a fine-grained insight about the architectures and reasoning that is required for solving the complex localization task.

\subsection{Architectures for Video Reasoning}
Several recent papers studied the effectiveness of CNN-based architectures for video action recognition. Many approaches use 3D convolutions for spatiotemporal feature learning \cite{carreira2017quo,tran2015learning} and separate the spatial and temporal modalities by adding optical flow as a second stream \cite{feichtenhofer2016spatiotemporal,simonyan2014two}. These models are computationally
expensive because 3D convolution kernels may be costly to compute. As a result, they may limited to sequence length to 20-30 frames \cite{carreira2017quo,tran2015learning}. In \cite{zhou2017temporalrelation} it was proposed to sparsely sample video frames to capture temporal relations in action recognition datasets. However, sparse sampling may be insufficient for long occlusion and containment sequences, which is the core of our OP focus.

Another strategy for temporal aggregation is to use recurrent architectures like LSTM \cite{hochreiter1997long}, connecting the underlying CNN output along the temporal dimension \cite{yue2015beyond}. \cite{gao2017video,song2017end,sharma2015action} combined LSTM with spatial attention, learning to attend to those parts of the video frame that are relevant for the task as the video progresses. In \secref{Experiments} we experiment with a spatial attention module, which learns to dynamically focus on relevant objects.

\subsubsection{Tracking with Object Occlusion.} A large body of work has been devoted to tracking objects \cite{mojtaba2019deep}. For objects under complex occlusion like carrying, early work studied tracking using classical techniques and without deep learning methods.
For instance,  \cite{huang2005tracking,papadourakis2010multiple} used the idea of object permanence to track objects through long-term occlusions. They located objects using adaptive appearance models, spatial distributions
and inter-occlusion relationships. 
In contrast, the approach presented in this paper focuses on a single deep differentiable model to learn motion reasoning end-to-end.
%from supervised data.
\cite{grabner2010tracking} succeeds to track occluded targets by learning how their movement is coupled with the movement of other visible objects. 
Unfortunately, the dataset studied here, CATER \cite{girdhar2019cater}, has weak object-object motion coupling by design. Specifically, when measuring the
correlation between the movement of the target and other object (as in \cite{grabner2010tracking}), we found that the correlation in 94\% of the videos was not statistically significant.

More recently, models based on Siamese neural network achieved SOTA results in object tracking~ \cite{Fan_2019,Li2018HighPV,Zhu_2018_ECCV}. Despite the power of these architectures, tracking highly-occluded objects is still challenging \cite{mojtaba2019deep}.
The tracker of \cite{Zhu_2018_ECCV}, DaSiamRPN, extends the region-proposal sub-network of \cite{Li2018HighPV}. It was designed for long-term tracking and handles full occlusion or out-of-view scenarios. DaSiamRPN was used as a baseline for the snitch localization task in CATER \cite{girdhar2019cater}, and we evaluated its performance for the OP problem in \secref{Experiments}.

\subsubsection{Containment.} Few recent studies explored the idea of containment relations.
{\cite{liang2018tracking} recovered incomplete object trajectories by reasoning about containment relations. \cite{ullman2019model} proposed an unsupervised model to categorize spatial relations, including containment between objects. The containment setup defined in these studies differs from the one defined here in that the contained object is always at least partially visible \cite{ullman2019model}, or 
the containment does not involve carrying \cite{liang2018tracking,ullman2019model}.}

% ==============================================
\section{The Learning Setup: Reasoning about Non-Visible Objects}
We next formally define the OP task and learning setup. 
We are given a set of videos $v_1,...,v_N$ where each frame $x^i_t$ in video $v_i$ is accompanied by the bounding box position $B^i_t$ of the target object as its label. The goal is to predict for each frame a bounding box $\hat{B}^i_t$ of the target object that is closest (in terms of  $L_1$ distance) to the ground-truth bounding box $B^i_t$. 

We define four localization tasks: (1) Localizing a visible object, which we define as an object that is at least partially visible. (2) Localizing an occluded object, which we define as an object that is \textit{fully} occluded by another object. (3) Localizing an object contained by another object, thus also completely non visible. (4) Localizing an object that is carried along the surface by a containing object. Thus in this case the target is moving while being completely non-visible. Together, these four tasks form a  localization task that we call \textbf{object-permanence localization task}, or OP.

In \secref{sect:abl}, we also study a semi-supervised learning setup, where at training time the location $B^i_t$ of the target is provided only in frames when it is visible. This would correspond to the case of a child learning object permanence without explicit feedback about where an object is located when it is hidden. 

It is instructive to note how the task we address here differs from the tasks of relation or reaction recognition \cite{krishna2017visual,lu2016visual,sadeghi2011recognition}.
In these tasks, models are trained to output an explicit label that captures the name of the interaction or relation (e.g., ``behind", ``carry"). In our task, the model aims to predict the location of the target (a regression problem), but it is not trained to name it explicitly (occluded, contained). While it is possible that the model creates some implicit representation describing the visibility type, this is not mandated by the loss or the architecture. 

\begin{figure}[t]
    \begin{center}
    \includegraphics[width=\textwidth,height=\textheight,keepaspectratio]{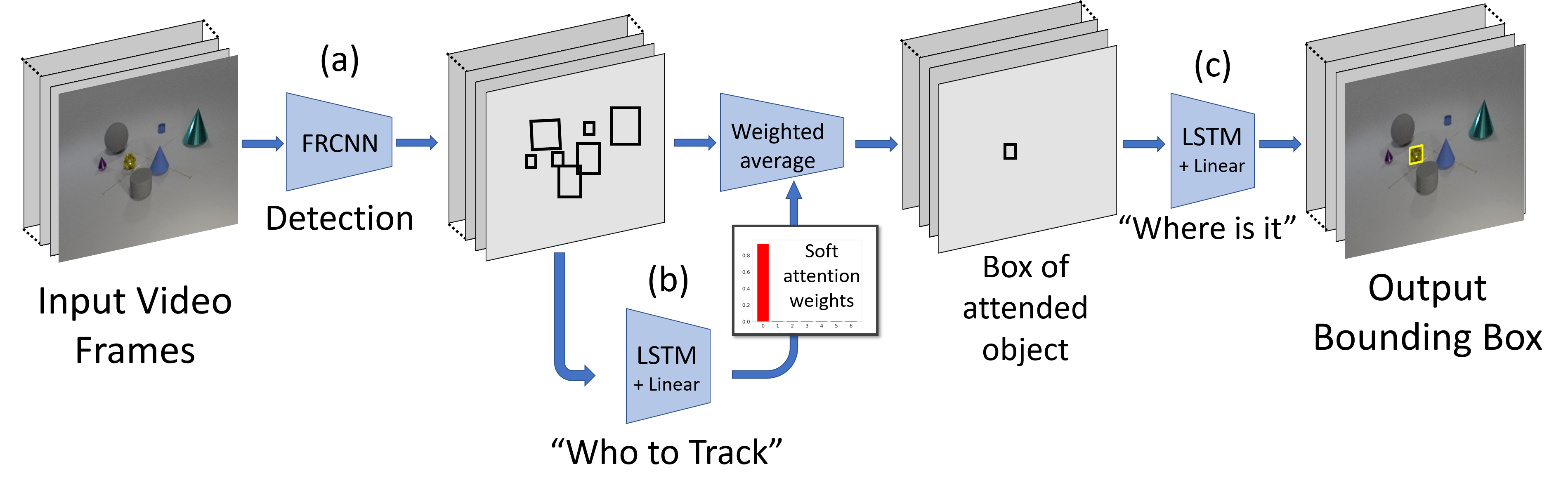}
    \caption{The architecture of \textit{Object Permanence network} (OPNet) consists of three components. (a) Perception module for detection. (b) Reasoning module for inferring which object to track in case the target is carried or contained. (c) A second reasoning module for occluded or visible targets, and for refining the exact location of the predicted target.
    \label{fig:opnet_arch}}
    \end{center}
\end{figure}

\newpage
% =================================
\section{Our Approach} 
\label{sec:model}
We describe a deep network architecture designed to address the four localization subtasks of the OP task. We refer to the architecture as OPNet. It contains three modules, that account for the perception and inference computations which facilitate OP (see \figref{fig:opnet_arch}).

\textit{Perception and detection module (\figref{fig:opnet_arch}a)}: A perception module, responsible for detecting and tracking visible objects. We incorporated a Faster R-CNN  \cite{ren2015faster} object detection model, fine-tuned on frames from our dataset, as the perception component of our model. 
After pre-training, we used the detector to output the bounding boxes together with identifiers of all objects in any given frame. Specifically, we represent a frame using a  $K\times 5$ matrix. Each row in the matrix represents an object using $5$ values: four values of the bounding box ($x_1, y_1, x_2, y_2$) and one {\em visibility bit}, which indicates whether the object is visible or not.
As the video progresses, we assign a unique row to each \emph{newly identified} object. If an object is not detected in a given frame, its corresponding information (assigned row) is set to zero. In practice, $K=15$ was the maximal number of objects in a single video in our dataset.
Notably, the videos in the dataset we used do not contain two identical objects, but we found that the detector sometimes mistakes one object for another.

\textit{``Who to track?" module (\figref{fig:opnet_arch}c)}: responsible for understanding which object is currently covering the target. This component consists of a single LSTM layer with a hidden dimension of 256 neurons and a linear projection matrix. After applying the LSTM to the object bounding boxes, we project its output to $K$ neurons, each representing a distinct object in the frame.  Finally we apply a softmax layer, resulting in a distribution over the objects in the frame. This distribution can be viewed as an attention mask focusing on the object that covers the target in this frame. Importantly, we do not provide explicit supervision to this attention mask (e.g., by explicitly ``telling the model" during training what is the correct attention mask). Rather, our only objective is the location of the target. The output of this module is $5$ numbers per frame. It is computed as the the weighted average over the $K\times 5$ outputs of the previous stage, weighted by the attention mask.

\textit{``Where is it" module (\figref{fig:opnet_arch}b)}: learns to predict the location of occluded targets. This final component consists of a second LSTM and a projection matrix. Using the output of the previous component, this component is responsible for predicting the target localization. It takes the output of the previous step ($5$ values per frame), feeds it into the LSTM and projects its output to four units, representing the predicted bounding box of the target for each frame.

\section{The LA-CATER Dataset}
To train models and evaluate their performance on the four OP subtasks defined above, we introduce a new set of annotations to the CATER dataset \cite{girdhar2019cater}. We refer to these as \textit{Localization Annotations} (LA-CATER). 

The CATER dataset consists of 5,500 videos generated programmatically using the Blender 3D engine. Each video is 10-second long (300 frames) and contains 5 to 10 objects. Each object is characterized by its shape (cube, sphere, cylinder and inverted cone), size (small, medium, large), material (shiny metal and matte rubber) and color (eight colors). Every video contains a golden small sphere referred to as ``the snitch", that is used as the target object which needs to be localized.

For the purpose of this study, we generated videos following a similar configuration to the one used by CATER, but we computed additional annotations during video generation.  Specifically, we augmented the CATER dataset with ground-truth bounding boxes locations of all objects. These annotations were programmatically extracted from the Blender engine, by projecting the internal 3D coordinates of objects are to the 2D pixel space.

We further annotated videos with detailed frame-level annotations. Each frame was labeled with one of four classes: visible, fully occluded, contained (i.e., covered, static and non-visible) and carried (i.e., covered, moving and non-visible). This classification of frames matches the four localization subtasks of the OP problem.
To compute these annotations, we computed the line-of sight from the camera position to determine if a target is occluded by another object, or occluding it.

\emph{LA-CATER} includes a total number of 14K videos split into train, dev and test datasets. See Table \ref{table:dataset} for a classification of video frames to each one of the localization subtasks across the dataset splits. Further details about dataset preparation are provided in appendix  \ref{sec:la_cater_prep}. 

\begin{table}[h]
    \setlength{\tabcolsep}{4pt}
    \begin{center}
    \textsc{
    \begin{tabular}{l|c|c|c|c|c|}
    \cline{2-6}
     & \begin{tabular}[c]{@{}c@{}}Number of \\ Frames\end{tabular} & \begin{tabular}[c]{@{}c@{}}Visible\end{tabular} & \begin{tabular}[c]{@{}c@{}}Occluded\end{tabular} & \begin{tabular}[c]{@{}c@{}}Contained\end{tabular} & \begin{tabular}[c]{@{}c@{}}Carried\end{tabular} \\ \hline
    \multicolumn{1}{|c|}{Train} & 9,300 & 63.00\% & 3.03\% & 29.43\%   & 4.54\%  \\ \hline
    \multicolumn{1}{|c|}{Dev}   & 3,327 & 63.27\% & 2.89\% & 29.19\%   & 4.65\%  \\ \hline
    \multicolumn{1}{|c|}{Test}  & 1,371 & 64.13\% & 3.07\% & 28.56\%   & 4.24\%  \\ \hline
    \end{tabular}
    }
    \end{center}
    \caption{Fraction of frames per type in the train, dev and test sets of LA-CATER. Occluded and carried target frames make up less than 8\% of the frames, but they present the most challenging prediction tasks.\label{table:dataset}}
\end{table}

% =====================================
\section{Experiments}
\label{Experiments}
We describe our experimental setup, compared methods and evaluation metrics. Implementation details are given in Appendix \ref{sec:implementation_details}.

\subsection{Baselines and Model Variants} \label{baseline_models}
We compare our proposed OPNet with six other architectures designed to solve the OP tasks. Since we are not aware of previously published unified architectures designed to solve all OP tasks at once, we used existing models as components in our baselines. All baseline models receive the predictions of the object detector (perception) component as their input.

\vspace{5pt}
\noindent\textit{(A) Programmed Models}. We evaluated two programmed models. These models are ``hard-coded" rather than learned. They are designed to reflect models that programmatically solve the reasoning task.
\begin{itemize}
    \item (1) \textbf{Detector + Tracker}. Using the detected location of the target, this method initiates a DaSiamRPN tracker \cite{Zhu_2018_ECCV} to track the target. Whenever the target is no longer visible, the tracker is re-initiated to track the object located in the last known location of the target.
    \item (2) \textbf{Detector + Heuristic}. When the target is not detected, the model switches from tracking the target to tracking the object located closest to last known location of the target. The model also employs an heuristic logic to adjust between the sizes of the current tracked object and the original target.
\end{itemize}

\noindent\textit{(B) Learned Models}. We evaluated four learned baselines with an increasing level of representation complexity.
\begin{itemize}
    \item (3) \textbf{OPNet}. The proposed model, as  presented in \secref{sec:model}.

    \item (4) \textbf{Baseline LSTM}. This model uses a single unidirectional LSTM layer with a hidden state of 512 neurons, operating on the temporal (frames) dimension. The input to the LSTM is the concatenation of the objects input representations. It is the simplest learned baseline as the input representation is not transformed non-linearly before being fed to the LSTM. 
    
    \item (5) \textbf{Non-Linear + LSTM}. This model augments the previous model and increases the complexity of the scene representation. The input representations are upsampled using a linear layer followed by a ReLU activation, resulting in a 256-dimensional vector representation for each object in the frame. These high-dimensional objects representations are concatenated and fed into the LSTM.
    
    \item (6) \textbf{Transformer + LSTM}. This model augments the previous baselines by introducing a much complex representations for objects in frame. We utilized a transformer encoder \cite{vaswani2017attention} after up-sampling the input representations, employing self attention between all the objects in a frame. We used a transformer encoder with 2 layers and 2 attention heads, yielding a single vector containing the target attended values. These attended values, which corresponds to each other object in the frame, are then fed into the LSTM.
    
    \item (7) \textbf{LSTM + MLP}. This model (\textit{Figure \ref{fig:opnet_arch}}) ablates the second LSTM module (c) in the model presented in \secref{sec:model}.
    \end{itemize}

\subsection{Evaluation Metric}
We evaluate model performance at a given frame $t$ by comparing the predicted target localization and the ground truth (GT) target localization. We use two metrics as follows. First, the intersection over union (IoU) metric.
\begin{equation}
    IoU_{t} =\frac{B^{GT}_{t} \cap B^{p}_{t}}{B^{GT}_{t} \cup B^{p}_{t}} \quad,
\end{equation}
where $B^{p}_{t}$ denotes the predicted bounding box for frame $t$ and $B^{GT}_{t}$ denotes the ground truth bounding box for frame $t$.

Second, we evaluate models using the mean average precision (MAP) metric. MAP is computed by employing an indicator function on each frame, determining whether the IoU value is greater than a predefined threshold, then averaging across frames in a single video and all the videos in the dataset.
\begin{equation}
AP = \frac{1}{n}\sum_{t=1}^{n}\mathbf{1_t} \quad\ \text{, where} \; \mathbf{1_t} = 
    \begin{cases}
    1 &IoU_{t} > IoU \; threshold \\
    0 & \rm{otherwise}
    \end{cases}
\end{equation}
\begin{equation}
    MAP = \frac{1}{N}\displaystyle\sum_{v=1}^{N}AP_{v} \quad.
\end{equation}
These per-frame metrics allow us to quantify the performance on each of the four OP subtasks separately.
 
\section{Results} 
\label{sec:Results}
We start with comparing OPNet with the baselines presented in  \secref{baseline_models}. We then provide more insights into the performance of the models by repeating the evaluations with \textit{``Perfect Perception"} in  \secref{sec:perfect_abl}.  \secref{sect:abl} describes a semi-supervised setting of training with visible frames only. Finally, in \secref{sec:cater_res} we compare OPNet with the models presented in the CATER paper on the original CATER data.

\begin{table}[!h]
\begin{center}
\setlength{\tabcolsep}{2.8pt}
\resizebox{\textwidth}{!}{%
\begin{tabular}{|l|c|c|c|c|c|}
\hline
\begin{tabular}[c]{@{}c@{}}Mean IoU$\pm$ SEM\end{tabular}              & \begin{tabular}[c]{@{}c@{}}Visible\end{tabular} & \begin{tabular}[c]{@{}c@{}}Occluded\end{tabular} & \begin{tabular}[c]{@{}c@{}}Contained\end{tabular} & \begin{tabular}[c]{@{}c@{}}Carried\end{tabular} & Overall           \\ \hline
\multicolumn{1}{|l|}{\textsc{Detector + Tracker}}   & 90.27 $\pm 0.13$ & 53.62 $\pm 0.58$ & 39.98 $\pm 0.38$ & 34.45 $\pm 0.40$ & 71.23 $\pm 0.51$ \\ \hline
\multicolumn{1}{|l|}{\textsc{Detector + Heuristic}} & 90.06 $\pm 0.14$ & 47.03 $\pm 0.73$ & 55.36 $\pm 0.53$ & 55.87 $\pm 0.59$ & 76.91 $\pm 0.43$ \\ \hline
\multicolumn{1}{|l|}{\textsc{Baseline LSTM}}        & 81.60 $\pm 0.19$ & 59.80 $\pm 0.61$ & 49.18 $\pm 0.64$ & 21.53 $\pm 0.40$ & 67.20 $\pm 0.53$ \\ \hline
\multicolumn{1}{|l|}{\textsc{Non-Linear + LSTM}}    & 88.25 $\pm 0.14$ & 70.14 $\pm 0.62$ & 55.66 $\pm 0.67$ & 24.58 $\pm 0.44$ & 73.53 $\pm 0.51$ \\ \hline
\multicolumn{1}{|l|}{\textsc{Transformer + LSTM}}   & \textbf{90.82} $\pm 0.14$ & \textbf{80.40} $\pm 0.61$ & 70.71 $\pm 0.78$ & 28.25 $\pm 0.45$ & 80.27 $\pm 0.50$ \\ \hline
\multicolumn{1}{|l|}{\textsc{OPNet (LSTM + MLP)}}            & 88.11 $\pm 0.16$ & 55.32 $\pm 0.85$ & 65.18 $\pm 0.89$ & \textbf{57.59} $\pm 0.85$ & 78.85 $\pm 0.52$ \\ \hline
\multicolumn{1}{|l|}{\textsc{OPNet (LSTM + LSTM)}}              & 88.89 $\pm 0.16$ & 78.83 $\pm 0.56$ & \textbf{76.79} $\pm 0.62$ & 56.04 $\pm 0.77$ & \textbf{81.94} $\pm 0.41$ \\ \hline
\end{tabular}%
}
\end{center}
\caption{Mean IoU performance of various models on the LA-CATER test data. ``$\pm$" denotes the standard error of the mean (SEM). OPNet performs consistently well across all subtasks. Also, on the contained and carried frames OPNet is significantly better than the other methods.
\label{table:od_results}}
\end{table}

\begin{figure}[!h]
\begin{center}
    \includegraphics[width=0.95\textwidth,height=\textheight,keepaspectratio]{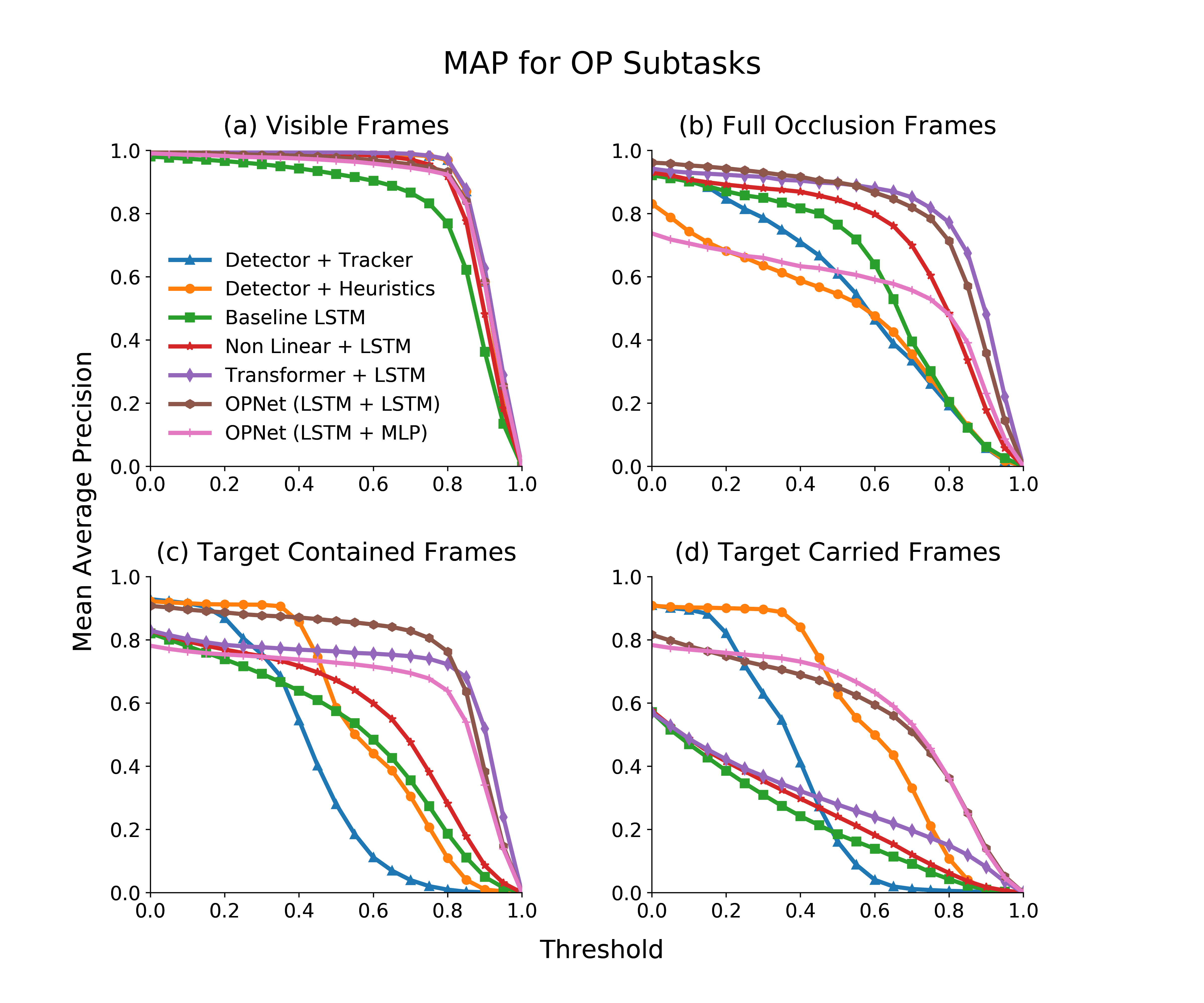}
    %\vspace{-10pt}
    \caption{Mean average precision (MAP) as a function of IoU thresholds. The two programmed models, Detector+Tracker (blue) and Detector+Heuristic (orange) perform well when the IoU threshold is low, providing a good coarse estimate of target location. OPNet performs well on all subtasks.\label{fig:map_od}}
\end{center}
\end{figure}

We first compare OPNet and the baselines presented in  \secref{baseline_models}.  Table \ref{table:od_results} shows IoU for all models in all four sub-tasks and  \figref{fig:map_od} presents the MAP accuracy of the models across different IoU thresholds.

It can be seen in Table \ref{table:od_results} that OPNet performs consistently well across all subtasks and outperforms all other models overall. On the visible and occluded frames performance is similar to other baselines. But on the contained and carried frames, OPNet is significantly better than the other methods. This is likely due to OPNet's explicit modeling of the object to be tracked.

Table \ref{table:od_results} also reports results for two variants of OPNet: OPNet (LSTM+MLP) and OPNet (LSTM+LSTM). The former is missing the second module (``Where is it" in \figref{fig:map_od}) which is meant to handle occlusion and indeed under-performs for occlusion frames (the ``occluded" and ``contained" subtasks). 
This highlights the importance of using the two LSTM modules in \figref{fig:map_od}.

\figref{fig:map_od} provides interesting insight into the behavior of the programmed models (namely Detector + Tracker and Detector + Heuristic). It can be seen that these models perform well when the IoU threshold is low. This reflects the fact that they have a good coarse estimate of where the target is, but fail to provide more accurate localization. On the other hand our OPNet model does well for accurate localization, presumably due to its learned ``Where is it" module.

\subsection{Reasoning with Perfect Perception} \label{sec:perfect_abl} 
The OPNet model contains an initial ``Perception" module that analyzes the frame pixels to get bounding boxes. Errors in this component will naturally propagate to the rest of the model and adversely affect results. Here we analyze the effect of the perception module by replacing it with ground truth bounding boxes and visibility bits. See Appendix \ref{sec:pp_annotation} for details on extracting ground-truth annotations. In this setup all errors reflect failure in the reasoning components of the models.

\begin{table}[!h]
    \begin{center}
    \resizebox{\textwidth}{!}{%
    \begin{tabular}{|l|c|c|c|c|c|}
    \hline
    Mean IoU $\pm$ SEM & Visible  & Occluded          & Contained         & Carried           & Overall           \\ \hline
    \textsc{DETECTOR + TRACKER}    & 90.27 $\pm 0.13$  & 53.62 $\pm 0.58$ & 39.98 $\pm 0.38$ & 34.45 $\pm 0.40$ & 71.23 $\pm 0.51$ \\ \hline
    \textsc{DETECTOR + HEURISTIC} & \textbf{95.59} $\pm 0.34$ & 30.40 $\pm 0.81$ & 59.81 $\pm 0.47$ & 59.33 $\pm 0.50$ & 81.24 $\pm 0.49$ \\ \hline
    \textsc{BASELINE LSTM}         & 75.22 $\pm 0.31$ & 50.52 $\pm 0.75$ & 45.10 $\pm 0.62$ & 19.12 $\pm 0.36$ & 61.41 $\pm 0.53$ \\ \hline
    \textsc{NON-LINEAR + LSTM}     & 88.63 $\pm 0.25$  & 65.73 $\pm 0.82$ & 58.77 $\pm 0.70$ & 23.89 $\pm 0.41$ & 74.53 $\pm 0.54$ \\ \hline
    \textsc{TRANSFORMER + LSTM}     & 93.99 $\pm 0.24$  & \textbf{81.31} $\pm 0.88$ & 75.75 $\pm 0.85$ & 28.01 $\pm 0.44$ & 83.78 $\pm 0.55$ \\ \hline
    \textsc{OPNet (LSTM + MLP)}    & 88.11 $\pm 0.16$  & 19.39 $\pm 0.60$ & 77.40 $\pm 0.68$ & \textbf{78.25} $\pm 0.65$ & 83.84 $\pm 0.48$ \\ \hline
    \textsc{OPNet (LSTM + LSTM)}   & 88.78 $\pm 0.25$  & 67.79 $\pm 0.69$ & \textbf{83.47} $\pm 0.47$ & 76.42 $\pm 0.66$ & \textbf{85.44} $\pm 0.38$ \\ \hline
    \end{tabular}%
    }
    \end{center}
    \caption{Mean IoU performance with the   \textit{Perfect Perception} setup. ``$\pm$" denotes the standard error of the mean (S.E.M.). Results are similar in nature to those with imperfect, detector-based, perception (Table \ref{table:od_results}). All models improve when using the ground-truth perception information. The subtask that improves the most with OPNet is the carried task.
    \label{table:perfect_per_results}}
\end{table}

\begin{figure}[!h]
    \begin{center}
    \includegraphics[width=0.95\textwidth,height=\textheight,keepaspectratio]{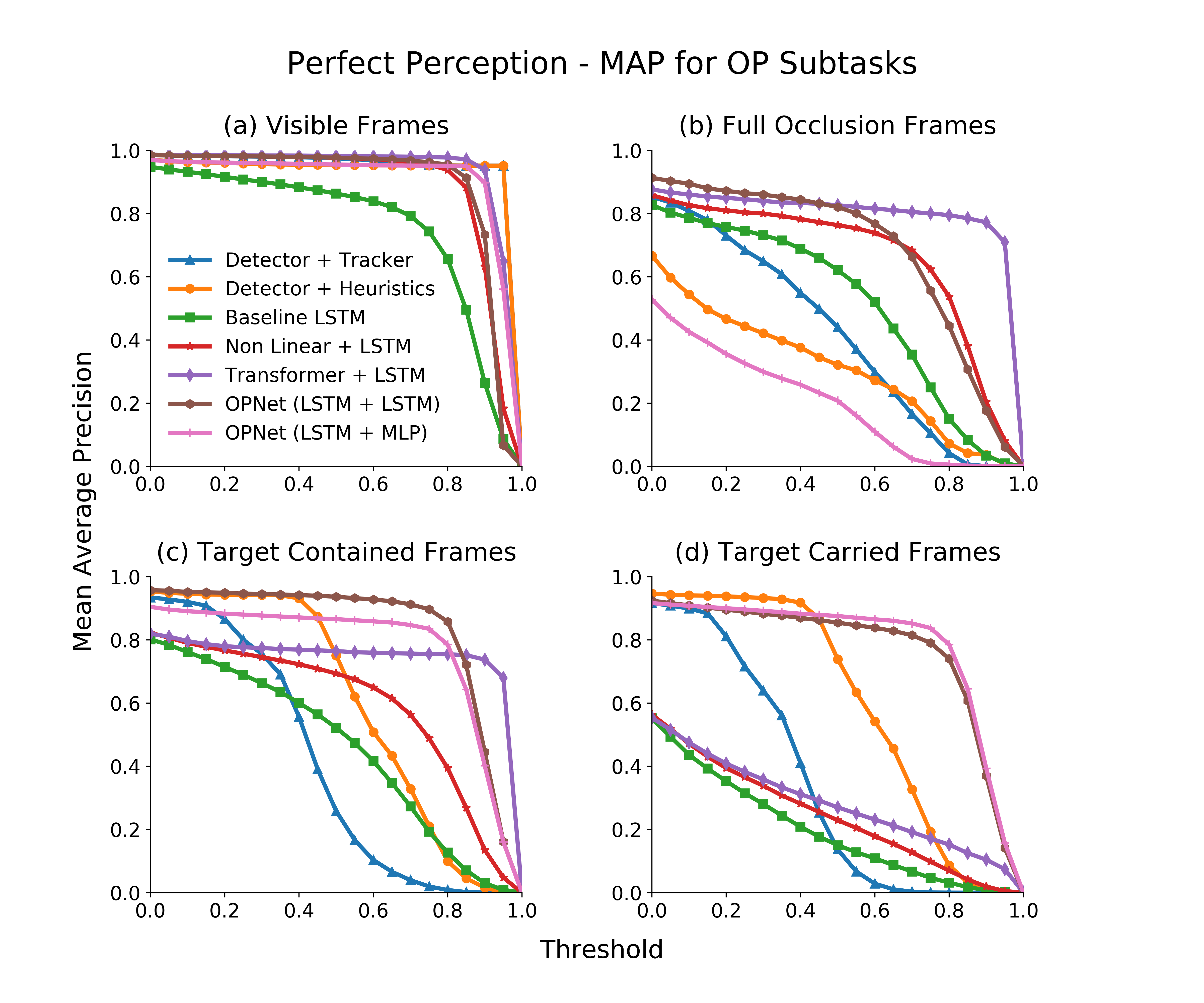}
    %\vspace{-10pt}
    \caption{Mean average precision (MAP) as a function of IoU thresholds for reasoning with Perfect Perception (\secref{sec:perfect_abl}). The most notable performance gain of OPNet (pink and brown curves) was with carried targets (subtask d).    \label{fig:map_pp}}
    \end{center}
\end{figure}

Table \ref{table:perfect_per_results} provides the IoU performance and Figure \ref{fig:map_pp} the MAP for all compared methods on all four subtasks. The results are similar to the previous results. When compared to the previous section (imperfect, detector-based, perception), the overall trend is the same, but all models improve when using the ground truth perception information. Interestingly, the subtask that improves the most from using ground truth boxes is the carried task. This makes sense, since it is the hardest subtask and the one that most relies on having the correct object locations per frame. 

\subsection{Learning only from Visible Frames}
\label{sect:abl}
We now examine a learning setup in which localization supervision is available only for frames where the target object is visible. This setup corresponds more naturally to the process by which people learn object permanence. For instance, imagine a child learning to track a carried (non visible) object for the first time and receiving a surprising feedback only when the object reappears in the scene.

In absence of any supervision when the target is non-visible, incorporating an extra auxiliary loss is needed to account for these frames. Towards this end, we incorporated an auxiliary \emph{consistency loss} that minimizes the change between predictions in consecutive frames. $  \mathcal{L}_{consistency} = \frac{1}{n} \sum_{t=1}^n \left\lVert b_{t} - b_{t-1}\right\rVert^2 $. The total loss is defined as an interpolation between the localization loss and the consistency loss, balancing their different scales: $ \mathcal{L} = \alpha \cdot \mathcal{L}_{localization} +  \beta \cdot \mathcal{L}_{consistency}$
Details on choosing the values of $\alpha$ and $\beta$ are provided in the supplementary.

\begin{table}[t]
    \begin{center}
    \resizebox{\textwidth}{!}{%
    \begin{tabular}{|l|c|c|c|c|c|}
    \hline
    \begin{tabular}[c]{@{}c@{}}Mean IoU\end{tabular}            & \begin{tabular}[c]{@{}c@{}}Visible\end{tabular} & \begin{tabular}[c]{@{}c@{}}Occluded\end{tabular} & \begin{tabular}[c]{@{}c@{}}Contained\end{tabular} & \begin{tabular}[c]{@{}c@{}}Carried\end{tabular} & Overall           \\ \hline
    \textsc{Baseline LSTM}& 88.61 $\pm 0.16$ & 80.39 $\pm 0.54$ & 68.35 $\pm 0.76$ & 27.39 $\pm 0.45$ & 78.09 $\pm 0.49$ \\ \hline
    \textsc{Non Linear + LSTM} & 89.30 $\pm 0.15$ & 82.49 $\pm 0.45$ & 67.25 $\pm 0.75$ & 27.34 $\pm 0.45$ & 78.15 $\pm 0.49$ \\ \hline
    \textsc{Transformer + LSTM} & 88.33 $\pm 0.15$ & \textbf{83.74} $\pm 0.44$ & \textbf{69.93} $\pm 0.77$ & \textbf{27.65} $\pm 0.54$ & 78.43 $\pm 0.49$ \\ \hline
    \textsc{OPNet (LSTM + MLP)} & 88.45 $\pm 0.17$ & 48.03 $\pm 0.82$ & 10.95 $\pm 0.51$ & 7.28 $\pm 0.30$ & 61.18 $\pm 0.69$ \\ \hline
    \textsc{OPNet (LSTM + LSTM)} & \textbf{88.95} $\pm 0.16$ & 81.84 $\pm 0.48$ & 69.01 $\pm 0.76$ & 27.50 $\pm 0.45$ & \textbf{78.50} $\pm 0.49$ \\ \hline
    \end{tabular}%
    }
    \end{center}
    \caption{IoU performance for the \textit{only visible supervision} setting. ``$\pm$" denote the standard error of the mean (S.E.M.). The models perform well when the target is visible, fully occluded or contained without movement, but not when the target is carried.
    \label{table:visible_only}}
\end{table}

Table \ref{table:visible_only} shows the mean IoU for this setup (compare with Table \ref{table:od_results}). The baselines perform well when the target is visible, fully occluded or contained without movement.
This phenomenon goes hand-in-hand with the inductive bias of the \emph{consistency loss}. Usually, to solve these subtasks, a model only needs to predict the last known target location. This explains why the OPNet (LSTM+MLP) model performs so poorly in this setup.

We note that the performance of non-OPNet models on the carried task is similar to that obtained using full supervision (see Table \ref{table:od_results}, \secref{sec:Results}) . This suggests that these models fail to use the supervision for the ``carried'' task, and further reinforces the observation that localizing carried object is highly challenging.

\subsection{Comparison with CATER Data} \label{sec:cater_res}
The original CATER paper \cite{girdhar2019cater} considered the ``snitch localization" task, aiming to localize the snitch at the last frame of the video, and formalized as a classification problem. The x-y plane was divided with a 6-by-6 grid, and the goal was to predict the correct cell of that grid. 

Here we report the performance of OPNet and relevant baselines evaluated on the exact setup as in \cite{girdhar2019cater}, to facilitate comparison between our models and the results reported there. Table \ref{table:cater_results}
shows the accuracy and $L_{1}$ distance metrics for this evaluation. OPNet significantly improves over all baselines from \cite{girdhar2019cater}. It cuts down the classification error from $40\%$ error down to $24\%$, and the $l_1$ distance from $1.2$ to $0.54$. 

\begin{table}[h]
    \begin{center}
    \begin{tabular}{|l|c|c|}
    \hline
    & Accuracy & $L_1$ Distance   \\ 
    Model & \small{(higher is better)}      & \small{(lower is better)}   \\ \hline
    \textsc{DaSiamRPN}             & 33.9 & 2.4 \\
    \textsc{TSN-RGB + LSTM}        & 25.6 & 2.6 \\
    \textsc{R3D + LSTM}            & 60.2 & 1.2 \\ \hline
    \textsc{OPNet (Ours)} & \textbf{74.8} & \textbf{0.54} \\ \hline
    \end{tabular}%
    \end{center}
    \caption{Classification accuracy on the CATER dataset using the metrics of \cite{girdhar2019cater}. OPNet significantly improves over all baselines for the ``snitch localization task".
    \label{table:cater_results}}
\end{table}

\begin{figure}[h!]
    \begin{center}
    (a) \hspace{155pt}
    (b) \hspace{155pt}~
    \includegraphics[width=0.495\textwidth]{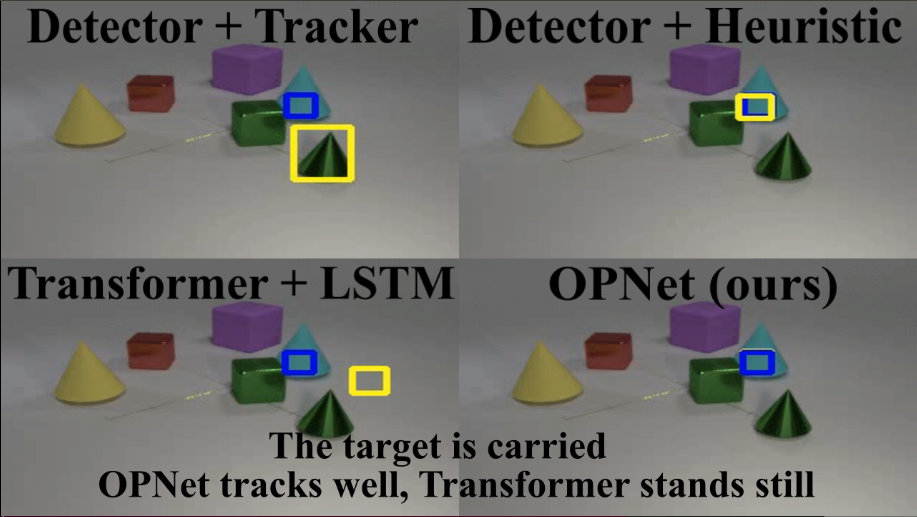}
    \includegraphics[width=0.495\textwidth]{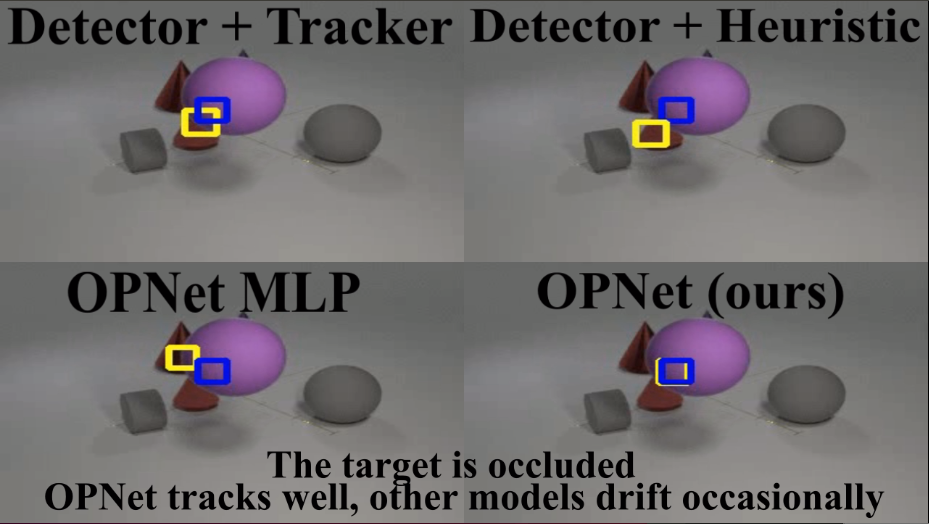}
    \caption{Screenshots from the model comparison video files. Blue boxes denote the ground truth location. Yellow boxes denote the predicted location. OPNet (ours) is at the bottom right panel. \textbf{(a)} The target is contained and then \textit{carried} by the blue cone and is captured successfully by OPNet. \textbf{(b)} The target is occluded by the red cone and purple ball. These occlusions confuse all baselines, while OPNet  localizes the target accurately.
    \label{fig:all_models_comparison}}
    \end{center}
\end{figure}

\begin{figure}[t!]
    \hspace{25pt} (a) \hspace{55pt}(b) \hspace{55pt}(c) \hspace{50pt} (d) \hspace{50pt} (e)
    \begin{center}
    \includegraphics[width=\textwidth,height=\textheight,keepaspectratio]{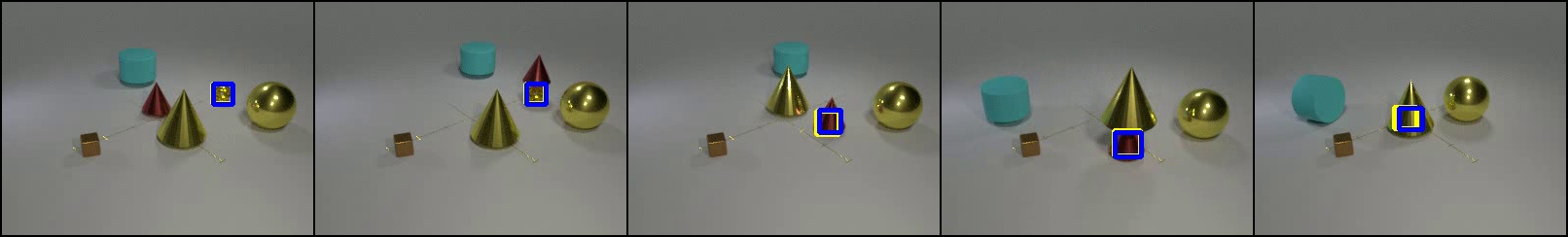}
    \includegraphics[width=\textwidth,height=\textheight,keepaspectratio]{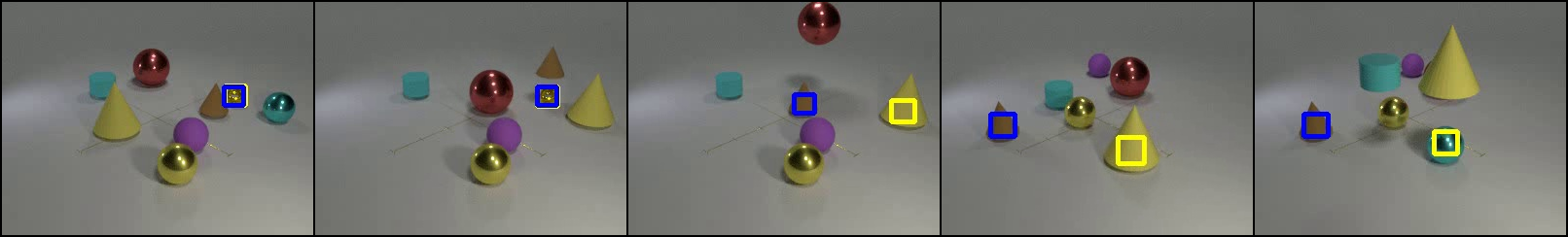}
    \caption{Examples of a success case (top row) and a failure case  (bottom row) for localizing a carried object. The blue box marks the ground-truth location. The yellow box marks the predicted location. \textit{Top} (a) The target object is visible; (b-c) The target becomes covered and carried by the orange cone; (d-e) The big golden cone covers and carries the orange cone, illustrating recursive containment. The target object is not visible, but OPNet successfully tracks it. \textit{Bottom} (c-d) OPNet accidentally switches to the wrong cone object (the yellow cone instead of the brown cone);
    (e) OPNet correctly finds when the yellow cone is picked up and switches to track the blue ball that was underneath.
    \label{fig:win_loss}}
    \end{center}
\end{figure}

\subsection{Qualitative Examples}
To gain insight into the successes and failures of our model, we now analyze specific examples. We provide two sets of examples to illustrate: (1) Comparison between baselines and variants over the same set of videos; (2) Wins and losses of our approach.

%Errors of our OPNet model typically correspond to cases where there
%are multiple objects that are candidates for covering the target and OPNet chooses the wrong one.

\textbf{Model Comparison}. We show two videos comparing OPNet with baselines and other variants. In both videos, four competing methods are applied to the same video scene. We recommend playing videos at a slow speed.

\begin{figure}[h!]
    \hspace{85pt}(a) \hspace{150pt} (b)
    \begin{center}
    \includegraphics[width=0.49\textwidth]{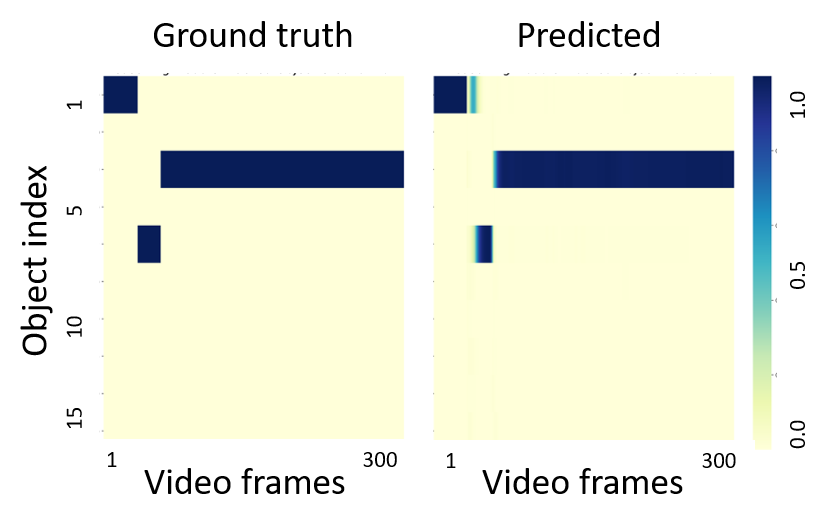}
     \includegraphics[width=0.49\textwidth]{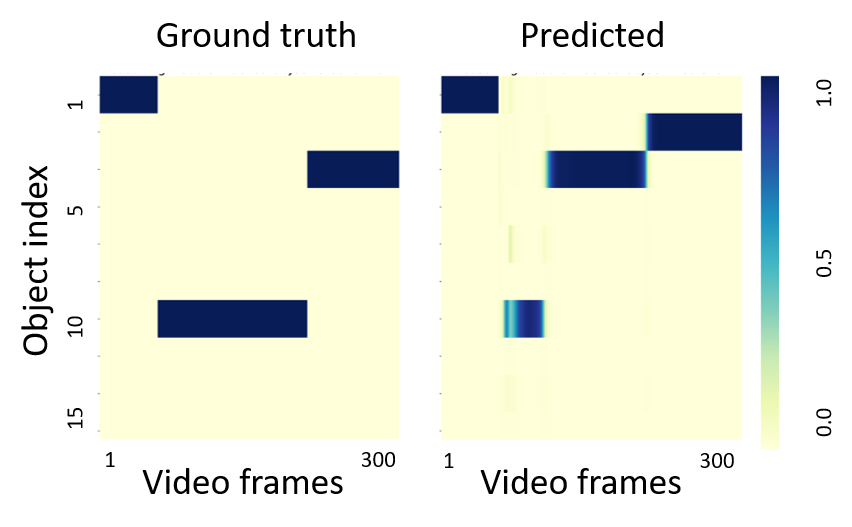}
    \caption{Switching attention across objects. In each pair of panels, each row traces the probability assigned to an object along the video in the ground truth (left) and predicted attention (right). (a) The system successfully switches attention from object 1 (target) when it is contained by object 6 and then carried by object 3. (b) After a successful switch from the object 1 to 10, the system incorrectly witches to object 3.
    \label{fig:heat_map}}
    \end{center}
\end{figure}

\begin{itemize}
    \item The first model comparison video (\href{https://youtu.be/TZgoxoKcGrE}{\textit{https://youtu.be/TZgoxoKcGrE}})  shows one visual scene analyzed by four methods.
    OPNet (ours) successfully localizes the target throughout the video. When the target is ``carried", the \textit{Transformer} model (bottom left) fails to switch and instead of tracking the carrying object it keeps predicting the last seen location of the target. The \textit{Tracker} model (top left) switches to a wrong object. The \textit{Heuristic} model (top right) successfully tracks the object containing the target, adjusting well to the target size. See \figref{fig:all_models_comparison}(a).
    \item  The second model comparison video  (\href{https://youtu.be/KoxbhgalazU}{\textit{https://youtu.be/KoxbhgalazU}}) shows a visual scene analyzed by four methods. In this video, the target is being occluded by multiple objects, including full occlusion, which makes it challenging to track. 
    The \textit{Tracker}, \textit{Heuristic} and \textit{OPNet MLP} models occasionally drift from the target when it is fully occluded by a large object.
    OPNet (ours) successfully localizes the target throughout the video. See \figref{fig:all_models_comparison}(b).
\end{itemize}
 
\textbf{Wins and Losses of OPNet}.
 We provide interesting examples of OPNet success and failures, adding insights into the behaviour and limitations of the OPNet model.

 \begin{itemize}
     \item The video \href{https://youtu.be/FnturB2Blw8}{\textit{https://youtu.be/FnturB2Blw8}} provides a ``win" example. It demonstrates the power of OPNet and its \textit{``who to track"} reasoning component. In the video, the model handles phases of recursive containment (``babushka"), which involve ``carrying". It suggests that OPNet learns an implicit representation of the object actions (pick-up, slide, contain etc.) even though it was not explicitly trained to do so. See Figure \ref{fig:win_loss} (top row)
     
     \item The video \href{https://youtu.be/qkdQSHLrGqI}{\textit{https://youtu.be/qkdQSHLrGqI}} illustrates a failure of our model. It shows an example where OPNet fails to switch between tracked objects when the target is ``carried". The model accidentally switches to a wrong cone object (the yellow cone) that already contains another object, not the target. Interestingly, OPNet properly identifies when the yellow cone is picked up and switches to track the blue ball that was contained by the yellow cone. It suggests that OPNet has implicitly learned the ``meaning" of actions performed by objects, without being explicitly trained to do so. See Figure \ref{fig:win_loss} (bottom row)
 \end{itemize}
 
Further insight may be provided by comparing the attention mask of the OPNet ``Who to Track" module and the ground-truth mask of the containing or carrying object. Figure \ref{fig:heat_map} compares these masks for success and failure cases. It can be seen that OPNet nicely tracks the correct object for most of the frames.

\section{Conclusion}
We considered the problem of localizing one target object in a highly dynamic scenes where the object can be occluded, contained or even carried away by another object. We name this task \textit{object permanence}, following the cognitive concept of a target object that is physically present in a scene but is occluded and carried in various ways. We presented an architecture called OPNet, whose components correspond to the natural perceptual and reasoning stages of solving OP. Specifically, it has a module that learns to switch attention to another object if it infers that the object contains or carries th target. Our empirical evaluation shows that these components are needed for improving accuracy in this task. 

Our results highlight a remaining gap between perfect perception and a pixel-based detector. It is expected that this gap may be even wider when applying OP to more complex natural videos in an open-world setting. It will be interesting to further improve detection architectures in order to reduce this gap. 

\subsection*{Acknowledgments} This study was funded by a grant to GC from the Israel Science Foundation (ISF 737/2018), and by an equipment grant to GC and Bar-Ilan University from the Israel Science Foundation (ISF 2332/18). AG received funding from the European Research Council (ERC) under the European Unions Horizon 2020 research and innovation program (grant ERC HOLI 819080).

\clearpage
\bibliographystyle{splncs04}
\bibliography{permanence}

% \bibliographystyleAppendix{splncs04}
% \bibliograpyAppendix{permanence}
\let\cleardoublepage\clearpage
% updated April 2002 by Antje Endemann
% Based on CVPR 07 and LNCS, with modifications by DAF, AZ and elle, 2008 and AA, 2010, and CC, 2011; TT, 2014; AAS, 2016; AAS, 2020

% \documentclass[runningheads]{llncs}
% \usepackage{graphicx}
% \usepackage{comment}
% \usepackage{amsmath,amssymb} % define this before the line numbering.
% \usepackage{acronym}

% \usepackage[resetlabels,labeled]{multibib}

% INITIAL SUBMISSION - The following two lines are NOT commented
% CAMERA READY - Comment OUT the following two lines
% \usepackage{ruler}
% \usepackage[width=122mm,left=12mm,paperwidth=146mm,height=193mm,top=12mm,paperheight=217mm]{geometry}
% \usepackage{tabularx, lipsum}
% \usepackage{hyperref}

% \usepackage{color}
% \begin{comment}
% \renewcommand\thefigure{S\arabic{figure}}    
% \renewcommand\thetable{S\arabic{table}}  
% \renewcommand\thesection{\Alph{section}}  
% \renewcommand{\theequation}{S\arabic{equation}}
% \setcounter{figure}{0}  
% \setcounter{table}{0}  
% \setcounter{section}{0} 
% \end{comment}

% \usepackage{xr}
% \makeatletter
% \newcommand*{\addFileDependency}[1]{% argument=file name and extension
%  \typeout{(#1)}
%  \@addtofilelist{#1}
%  \IfFileExists{#1}{}{\typeout{No file #1.}}
% }
% \makeatother

% \newcommand*{\myexternaldocument}[1]{%
%     \externaldocument{#1}%
%     \addFileDependency{#1.tex}%
%     % \addFileDependency{#1.aux}%
% }
% \myexternaldocument{main}

% \acrodef{OP}{Object Permanence}
\definecolor{orange}{RGB}{255,127,0}
\definecolor{green}{RGB}{0,127,0}

\renewcommand\thefigure{S\arabic{figure}}    
\renewcommand\thetable{S\arabic{table}}  
\renewcommand\thesection{\Alph{section}}  
\renewcommand{\theequation}{S\arabic{equation}}

% \newcommand{\galch}[1]{{\color{blue}{\bf[GC:} #1{\bf]}}}
% \newcommand{\amirg}[1]{{\color{orange}{\bf[AG:} #1{\bf]}}}
% \newcommand{\amir}[1]{{\color{orange}{#1}}}
% \newcommand{\gal}[1]{{\color{blue}{#1}}}
% \newcommand{\avivsh}[1]{{\color{red}{\bf[AS:} #1{\bf]}}}
% \newcommand{\aviv}[1]{{\color{red}{#1}}}
% \newcommand{\ofrik}[1]{{\color{green}{\bf[OK:} #1{\bf]}}}
% \newcommand{\ofri}[1]{{\color{green}{#1}}}
% \newcommand{\secref}[1]{Section \ref{#1}}
% \newcommand{\figref}[1]{Figure \ref{#1}}
% \renewcommand{\eqref}[1]{Equation \ref{#1}}

% \begin{document}

% \linenumbers
\pagestyle{headings}
\mainmatter
% \def\ECCVSubNumber{2481}  % Insert your submission number here

% \title{Learning Object Permanence from Video} % Replace with your title

%\begin{comment}
% \titlerunning{ECCV-20 submission ID \ECCVSubNumber} 
% \authorrunning{ECCV-20 submission ID \ECCVSubNumber} 
% \author{Anonymous ECCV submission}
% \institute{Paper ID \ECCVSubNumber}
%\end{comment}
%******************

\begin{comment}
\titlerunning{Learning Object permanence}
\author{First Author\inst{1}\orcidID{0000-1111-2222-3333} \and
Second Author\inst{2,3}\orcidID{1111-2222-3333-4444} \and
Amir Globerson\inst{2,3}\orcidID{1111-2222-3333-4444} \and
Gal Chechik\inst{3}\orcidID{2222--3333-4444-5555}}
%
\authorrunning{F. Author et al.}
\institute{Bar-Ilan University, Ramat-Gan, Israel \and
nvidia research, Israel
\email{Gal.Chechik@biu.ac.il}\\
\url{http://chechiklab.biu.ac.il} \and
ABC Institute, Rupert-Karls-University Heidelberg, Heidelberg, Germany\\
\email{\{abc,lncs\}@uni-heidelberg.de}}
\end{comment}
%******************
\author{}
\institute{}
\title{Supplementary Material} 
% \titlerunning{ECCV-20 submission ID \ECCVSubNumber} 
% \authorrunning{ECCV-20 submission ID \ECCVSubNumber} 
% \author{Anonymous ECCV submission}
% \institute{Paper ID \ECCVSubNumber}
\maketitle

\appendix

\section{Erorr Analysis across the Video Corpus}
Videos in our dataset vary substantially in terms of what OP tasks they involve. This has a large effect over localization accuracy, because it is much harder to localize a carried target  than a visible one. To gain more insight into the performance of the leading models, we compare the localization IoU on a video-by-video basis.

\begin{figure}[]
    \hspace{90pt}(a) \hspace{150pt} (b)
    \begin{center}
    \includegraphics[width=0.495\textwidth]{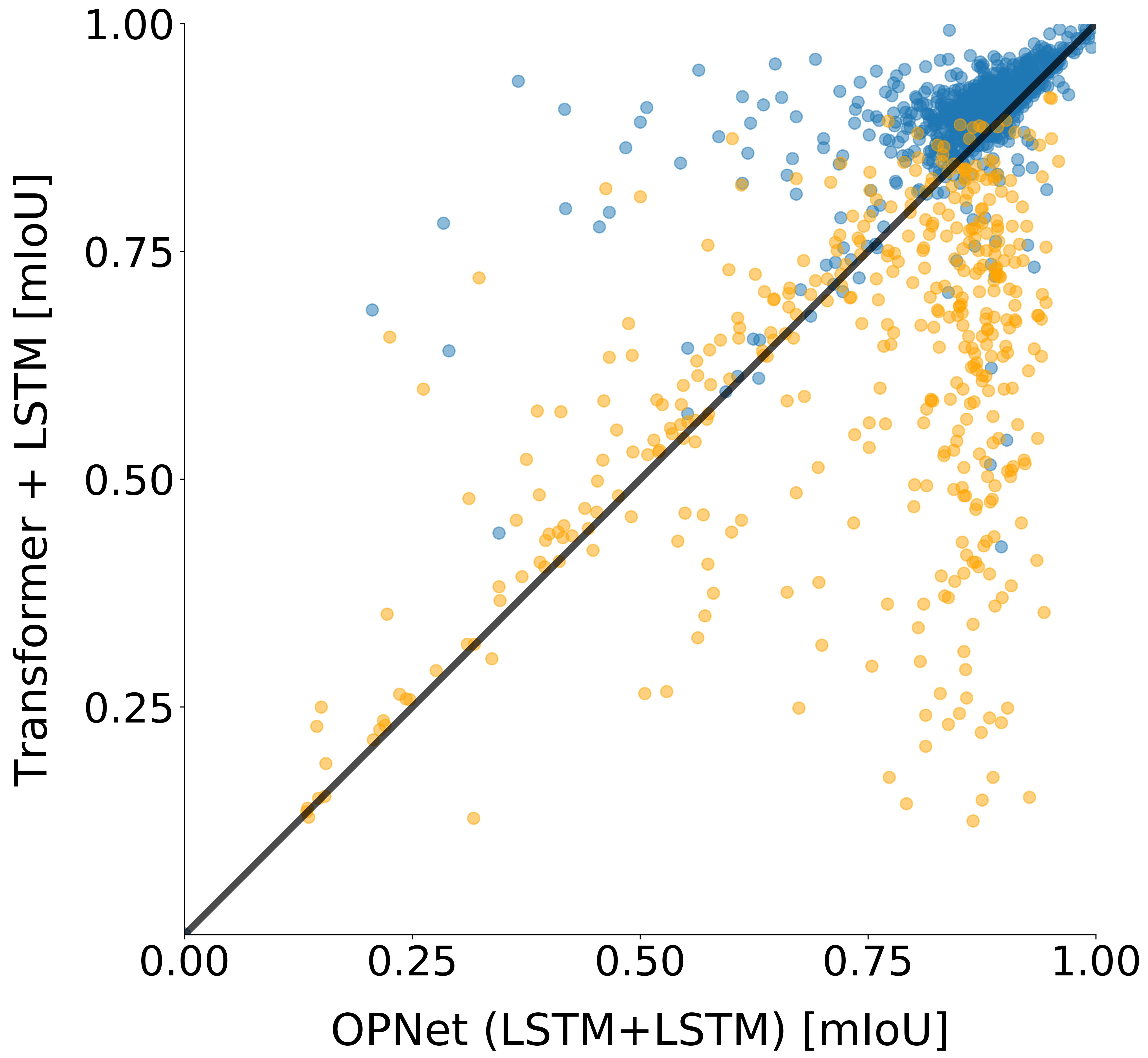}
    \includegraphics[width=0.495\textwidth]{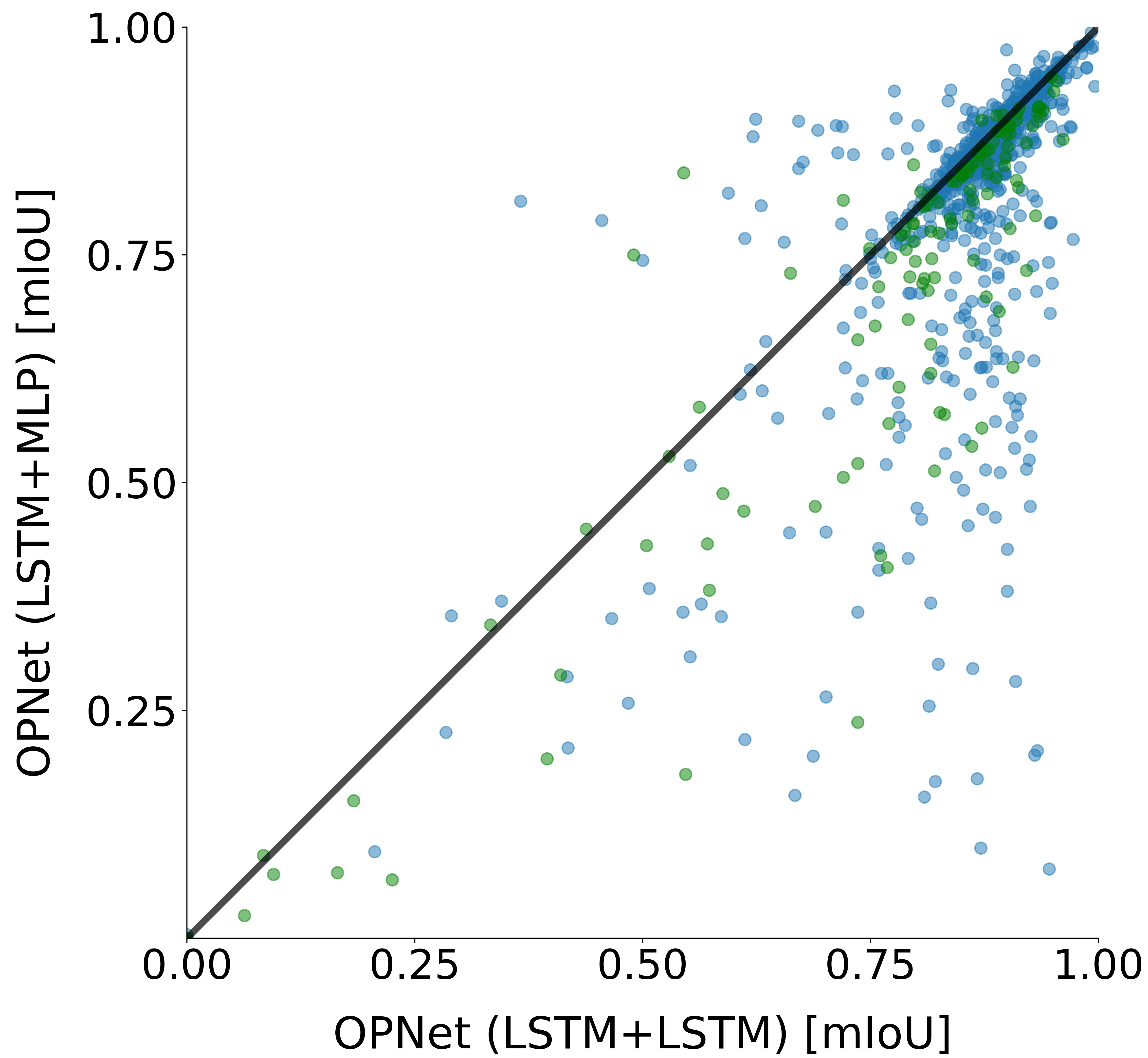}
    \caption{Sample-by-samples comparison of OPNet with two strong baselines. Each point represents the IoU of a video from the test set, achieved by OPNet and a baseline. \textbf{(a)} Videos with more than 7\% \textit{carried} frames are colored in orange. \textbf{(b)} Videos with more than 7\% \textit{occlusion} frames are colored in green. Points in the lower part corresponds to videos in which OPNet is superior.}
    \label{fig:overall_iou_model_comparison}
    \end{center}
\end{figure}

Figure \ref{fig:overall_iou_model_comparison} depicts per-video IoU of OPNet and two other strong baselines. Each point corresponds to one video and the color reflects the type of frames in that video. Figure \ref{fig:overall_iou_model_comparison}(a) shows how OPNet outperforms Transformer on videos including \textit{carried} frames (colored in orange). Clearly, videos with carried frames are clustered in the lower half of the figure, where OPNet is superior.

Similarly, \figref{fig:overall_iou_model_comparison}(b) compares OPNet with the OPNet (LSTM + MLP) baseline, which contains only the first reasoning component (see Our Approach section). It shows that OPNet outperforms the baseline on videos including a high number of \textit{occlusion} frames (colored in green). It also emphasizes that for most videos, OPNet is superior, as illustrated by the great number of points in the lower half of the figure.

\section{Implementation Details}\label{sec:implementation_details}
We trained OPNet and baseline variants using $L_1$ loss optimized using Adam optimizer with $\beta_1=0.9$, $\beta_2=0.999$, $\varepsilon=1e-08$, and using a batch size of 16. We initialized the learning rate to $0.001$ and employed a learning rate decay policy, which reduced the learning rate by a factor of 0.8 every 3 epochs without loss improvement. We tuned all hyperparameters using the validation set. We experimented with using a higher initial learning rate of $1e-2$, but it turned out to be too noisy for the relatively small loss induced by the $L_1$ loss. We also tried lower learning rate ($1e-4$), but it did not converge to a good minimum.

The model was trained for 160 epochs, which we verified via manual inspection to be sufficient for convergence of all models. Early stopping was based on the validation-set mean IoU.

For comparisons with CATER \cite{girdhar2019cater} (Table \ref{table:cater_results} of the main paper), we used the accuracy values reported in their paper. 

For the \textit{learning only from visible frames} setup (Section \ref{sect:abl} and Table \ref{table:visible_only} of the main paper) we used the values $\alpha = 1$ and $\beta = 0.5$. We used these values to normalize the different scales of $\mathcal{L}_{localization}$ and $\mathcal{L}_{consistency}$. We verified via manual inspection that (1) for the first 60-70 epochs the loss component $\mathcal{L}_{localization}$ is significantly greater than the loss component $\mathcal{L}_{consistency}$. Thus, in this phase the model improves its prediction when the target is visible; (2) After 60-70 epochs the two loss components have the same scale. Thus, in this phase the model improves its prediction also when the target is not-visible.

\section{LA-CATER Dataset Preparation} \label{sec:la_cater_prep} Our new LA-CATER dataset augments the CATER dataset \cite{girdhar2019cater} with ground-truth locations of all objects and with detailed frame level annotations. Also, instead of using the videos released by CATER we
generated new videos using their configuration, and 
expanded their code to add ground-truth locations and frame-level annotations. 

We now describe how we classify frames into the four corresponding OP subtasks. The CATER dataset annotates each frame with the \textit{actions} occurring for each object in that frame. These actions are defined as follows:
\begin{itemize}
    \item \textit{Slide}.
    Object changes its position by sliding on the XY-plane. 
    \item \textit{Pick-Place}. 
    Object is picked up in the air along the Z-axis, moved to a new location and placed down.
    \item \textit{Contain}. 
    A special action performed by cones only, in which cone execute \textit{Pick-Place} action and positioned on top of another object.
\end{itemize}

\begin{itemize}
\item \textbf{Contained Frames}.
 We classify a frame as \textit{Contained} when the target is contained by a cone. Explicitly, a frame is classified as \textit{Contained} when it is annotated with the ``contain" action in CATER, with a cone marked as the containing object and the target marked as the contained object. A frame with recursive containment, namely, a containing cone is itself contained by another cone, is also considered to be  a \textit{contained} frame. Frames are marked as contained from the moment the target is covered and until the containing object is picked up as part of \textit{pick-place} action.

\item
\textbf{Carried Frames}. We mark a frame as \textit{Carried} when the target is \textit{contained} by a cone (its action is marked in CATER as contained) and \textit{slides} along with it. Frames are marked as \textit{carried} from the beginning of the \textit{slide} action until the end of the \textit{slide} action. Thus, only frames corresponding to the \textit{slide} action are marked as \textit{carried}.

\item
\textbf{Occluded Frames}. For frame $t$, we define the \textit{occlusion rate (OR)} of object $x$ by object $y$ as
\begin{equation}
{OR^{x}_{t}(y)} = 
    \begin{cases}
    \frac{Area^{x}_{t} \cap Area^{y}_{t}}{Area^{x}_{t}} &Area^{x}_{t} \leq Area^{y}_{t} \\
    \quad \quad 0 & Otherwise %Area^{x}_{t} > Area^{y}_{t}
    \end{cases}
\end{equation}
Where $Area^{x}_{t}$ is the area of object $x$ in frame $t$.
% $loc^{x}_{t}$ and $loc^{CA}_{t}$ are the locations of object $x$ and the camera in frame $t$ respectively (represented by 3D coordinates).

We define the \textit{distance from camera (DC)} of object $x$
\begin{equation}
    DC^{x}_{t} = \left\lVert loc^{x}_{t} - loc^{CA}_{t}\right\rVert^2 \quad
\end{equation}
$loc_t^x$, $loc_t^{CA}$ denote the 3D coordinates location of object $x$ and the camera in frame $t$ respectively.

We define an indicator for a \textit{fully occluded (FO)} object:
\begin{equation}
    {FO^{x}_{t}} = 
    \begin{cases}
    1 &\exists \;\; y \; \; s.t \; \; OR^{x}_{t}(y) = 1 \; \; \text{and} \; \; DC^{x}_{t} \geq DC^{y}_{t} \\
    0 & \quad \quad \quad \quad Otherwise
    \end{cases}
\label{eq:fo}
\end{equation}

We then mark frame $t$ as \textit{Occluded} when the target is fully occluded by another object. e.g $FO^{target}_{t} = 1$

\item
\textbf{Visible Frames}. Finally, we define frame as \textit{Visible} when the target is not \textit{Contained}, \textit{Carried} or \textit{Occluded}. Thus, the target needs to be only partially visible to be considered as \textit{visible}. For instance, the target is still considered \textit{visible} when it is 20\% occluded (e.g $\exists \;\; y \; \; s.t \; \; OR^{x}_{t}(y) = 0.2$)

% Thus, the target needs to be only partially visible \galch{Be specific} to be considered as visible.
\end{itemize}

\section{Annotating Frames in Perfect Perception}
\label{sec:pp_annotation}
For the perfect-perception setup, we extend the definition of \textit{fully occluded (FO)} objects from Eq \ref{eq:fo}.
We define an object to be \textit{partially occluded (PO)} with respect to the rate $p$ as follows:
\begin{equation}
{PO^{x}_{t}(p)} = 
    \begin{cases}
    1 &\exists\;\; y \; \; s.t \; \; OR^{x}_{t}(y) \geq p \; \; \text{and} \; \; DC^{x}_{t} \geq DC^{y}_{t} \\
    0 & \quad \quad \quad \quad Otherwise
    \end{cases}
\label{eq:po}
\end{equation}
We say that object $x$ is non-visible in frame $t$ with respect to $p$ if $PO^{x}_{t}(p) = 1$. We use the value $p = 0.7$ to decide which objects are non-visible. Contained objects are defined as non-visible, regardless of their $PO$ value. 

Objects are represented by a 5-coordinate vector, containing 4 bounding box coordinates in $(x_{1},y_{1},x_{2},y_{2})$ format and an additional visibility bit. Visible objects are represented by their ground-truth bounding boxes and a turned-on visibility bit. Non-visible objects are represented by a four-zeros bounding box coordinates and a turned-off visibility bit.
% \end{document}

\end{document}